\def\year{2020}\relax
\documentclass[letterpaper]{article} 
\usepackage{aaai20}  
\usepackage{times}  
\usepackage{helvet} 
\usepackage{courier}  
\usepackage[hyphens]{url}  
\usepackage{graphicx} 
\usepackage{graphicx}  
\usepackage{amssymb}
\usepackage{amsmath}
\usepackage{mathtools}
\title{Shape-Oriented Convolution Neural Network for Point Cloud Analysis}
\author{Chaoyi Zhang,\textsuperscript{\rm 1} Yang Song,\textsuperscript{\rm 2} Lina Yao,\textsuperscript{\rm 2} Weidong Cai\textsuperscript{\rm 1}\\
\textsuperscript{\rm 1} School of Computer Science, University of Sydney, Australia\\
\textsuperscript{\rm 2} School of Computer Science and Engineering, University of New South Wales, Australia\\
$\{$chaoyi.zhang, tom.cai$\}$@sydney.edu.au, $\{$yang.song1, lina.yao$\}$@unsw.edu.au
}

\begin{document}

\maketitle

\begin{abstract}
Point cloud is a principal data structure adopted for 3D geometric information encoding. Unlike other conventional visual data, such as images and videos, these irregular points describe the complex shape features of 3D objects, which makes shape feature learning an essential component of point cloud analysis. To this end, a shape-oriented message passing scheme dubbed \textit{ShapeConv} is proposed to focus on the representation learning of the underlying shape formed by each local neighboring point. Despite this \textit{intra-shape relationship} learning, \textit{ShapeConv} is also designed to incorporate the contextual effects from the \textit{inter-shape relationship} through capturing the long-ranged dependencies between local underlying shapes. This shape-oriented operator is stacked into our hierarchical learning architecture, namely Shape-Oriented Convolutional Neural Network (SOCNN), developed for point cloud analysis. Extensive experiments have been performed to evaluate its significance in the tasks of point cloud classification and part segmentation.
\end{abstract}


\section{Introduction}
As a principal data structure adopted for 3D geometric information encoding, point cloud has been widely used in several practical applications, such as self-driving cars and computer graphics. Although geometrical and topological information can be well-described as the raw point coordinates in point cloud data, the further analysis step of these irregular points can be quite challenging, as the local underlying shapes may not be modeled or recognized appropriately. Following the significant success recently achieved by Convolution Neural Networks (CNNs) on regular-formatted visual data, such as images and videos, several attempts have been made to transform raw point cloud data to either 3D volumetric representations or a collection of 2D views, so that it can be handled directly by the CNNs defined on regular grids. 

However, these approaches all have their own drawbacks and limitations. For the voxelization-based approaches, the potential information loss occurred in the voxelization stage may irrupt the object shapes by introducing quantization error. For the view-based methods, although an accurate classification or a descent segmentation can be reached, it requires a large number of view images collected from different angles, to make sure the generated 2D projections contain enough discriminative representations of the 3D objects for further point cloud analysis. Thus, a more shape-oriented geometric learning approach needs to be developed, which is not only expected to be able to manipulate the point cloud data directly, but is also designed to be capable of understanding the discriminative information of each local underlying shape, by modeling the shape-oriented contextual information encoded in the point cloud.

PointNet~\cite{pointnet} was the first deep learning based approach to manipulate the point cloud data directly, learning the pointwise features independently and outputting a global shape signature from the symmetric aggregation function applied to these pointwise features. Following the PointNet, the entire community started to pay more attention to local point cloud structure modeling, which was neglected completely by PointNet. DGCNN~\cite{dgcnn} could be seen as the next milestone, as they generally define \textit{EdgeConv} as:
\begin{equation}
    \mathbf{x}_{i}^{k}=\underset {j : (i, j) \in \mathcal{E}} {\square} h_{\Theta}\left(\mathbf{x}_{i}^{k-1}, \mathbf{x}_{j}^{k-1}\right),
\end{equation} where $k$ is the network layer index, $X_i$ and $X_j$ indicate the central point and the neighboring points, $\mathcal{E}$ denotes the local graph dynamically constructed among these points, and $\square$ is the feature aggregation function covering the entire local neighbourhood. As one of their main contributions, the $EdgeConv$ proposed helps to reformulate the learning process of the geometric information as the aggregation outcomes from the edge features computed pairwisely. It indicates that to update the pointwise feature $X_i$, the dynamic local graphs are firstly constructed using k-nearest neighbors searching technique, then centralized connections can be built between the centroid point and its neighbours, and the edge features $h_{\Theta}(\cdot, \cdot)$ are computed for the updating of the pointwise feature for the centroid points.
The definition of \textit{EdgeConv} reveals the idea that most of the authors claim that the local underlying shapes can be learnt as the aggregated pairwise interactions between the centroid point and its neighbor points. 

Unlike their approaches, we propose the shape-oriented convolution (\textit{ShapeConv}) to link all points forming the local shape to enhance their individual interactions and compute the moment point of this densely-connected local graph. After that, we show that the overall interaction between each point and the local shape can be simplified as the pointwise interaction placed between each point and the moment point. Furthermore, we study the contextual information encoding, via defining and modeling two shape-oriented relationships for point clouds, namely the intra-shape relationship and the inter-shape relationship. By incorporating this contextual information, our \textit{ShapeConv} module proposed is capable of performing several advanced point cloud analyses. To this end, we stack \textit{ShapeConv} into our shape-oriented convolution neural network (\textit{SOCNN}) and evaluate its significance in the tasks of point cloud classification and part segmentation.

\section{Related Work}
In this section, we will mainly review the previous works performed toward point cloud analysis, from three perspectives: the voxelization-based approaches, the view-based approaches, and the geometric learning approaches.
\subsubsection{Voxelization-based Approaches}
Voxelization is a particular kind of transformation, which takes irregular point cloud data as input and transforms them into several volumetric objects represented under a regular 3D coordinate system. Benefiting from the new volumetric representations, point cloud data can thus be processed conveniently by the 3D Convolution operators defined on regular 3D grids. Hence, several advanced point cloud analyses, such as the classification and segmentation tasks, could be easily achieved by the CNN, whose discriminative capabilities have been broadly evaluated in the computer vision community~\cite{imagenet,resnet,densenet,senet}. However, the performance of these voxelization-based approaches~\cite{modelnet40,voxnet} to point cloud analysis would be largely constrained by the critical shape information loss during the quantization step of the voxelization procedure. Although several subvolume-related works have been proposed to alleviate this spatial information loss~\cite{kdnet,ocnn,octnet}, their resulting subdivision representations can still be suffering from the potential quantization error, compared to other approaches modeling the local underlying shapes directly.
\subsubsection{View-based Approaches}
Rather than representing point cloud data from 3D regular grids, like the voxelization-based approaches mentioned above, view-based approaches are focused on recognizing and analyzing the point cloud data, from collections of 2D views. Due to the promising results achieved by 2D CNNs, view-based approaches can achieve excellent outcomes for point cloud analysis~\cite{multiview,multiview2,deepshape}. However, their side effects should be taken into consideration as well. That is, to reach an accurate classification or a descent segmentation, it requires a large number of views to be used for model training process. Meanwhile, to ensure the information encoded in the views are discriminative enough for the shape recognition, these views should be captured from as many different angles as possible~\cite{groupview}, which results in a long rendering time needed for the data sampling stage.
\subsubsection{Geometric Learning Approaches}
In contrast to the approaches above, geometric learning approaches have been developed recently, with the aim of processing 3D point cloud directly. PointNet~\cite{pointnet} and PointNet++~\cite{pointnet2} were the first to propose this manner of direct point cloud data manipulation, which could fundamentally solve the quantization error problem caused by the voxelization-based approaches and requires significantly less sampling data than the view-based approaches. Among all geometric learning approaches developed~\cite{pointnet,pointnet2,monet,PCNN,rscnn,pointweb,pointwise}, DGCNN~\cite{dgcnn} can be seen as a domain milestone, due to their general definition of \textit{EdgeConv}. As clearly illustrated in their work, most of the geometric learning approaches, including PointNet and PointNet++, can be considered as special instances from the general \textit{EdgeConv} family. However, unlike these instances, we revisit the geometric learning of point cloud data from a shape-oriented level and, based on this, we define two shape-oriented relationships occurring in point cloud data, which encode the global context information and local context information respectively. The pointwise features can therefore be updated by incorporating the contextual effects caused by the intra-shape relationships and inter-shape relationships captured for point cloud data.

\section{Method}
Let $\mathbb{P} = \{P_1, \ P_2, \ ..., \ P_N\}$ and $\mathbb{X} = \{X_1, \ X_2, \ ..., \ X_N\}$ denote the point cloud to be analyzed and its pointwise features, respectively, where $N$ is the number of points sampled and $C$ is the number of channels of each point feature, such that $X_i \in \mathbb{R}^{C}$. For a given sampled point $P_i$, we represent its neighborhood as $\mathcal{N}(P_i)$, and the local shape formed by this neighbourhood area as $\mathcal{S}_{\mathcal{N}(P_i)}$. 

\subsection{ShapeConv: Shape-Orientated Convolution}
We argue that the pointwise feature $X_i$ should be capable of encoding the characterization of local shape $\mathcal{S}_{\mathcal{N}(P_i)}$. Meanwhile, considering the global influences between each local shape, $X_i$ should also be designed to capture the long-ranged dependencies between $\mathcal{S}_{\mathcal{N}(P_i)}$ and $\mathcal{S}_{others}$ including all other local shapes. 

\begin{figure*}[!t]
\centering
\includegraphics[width=\textwidth]{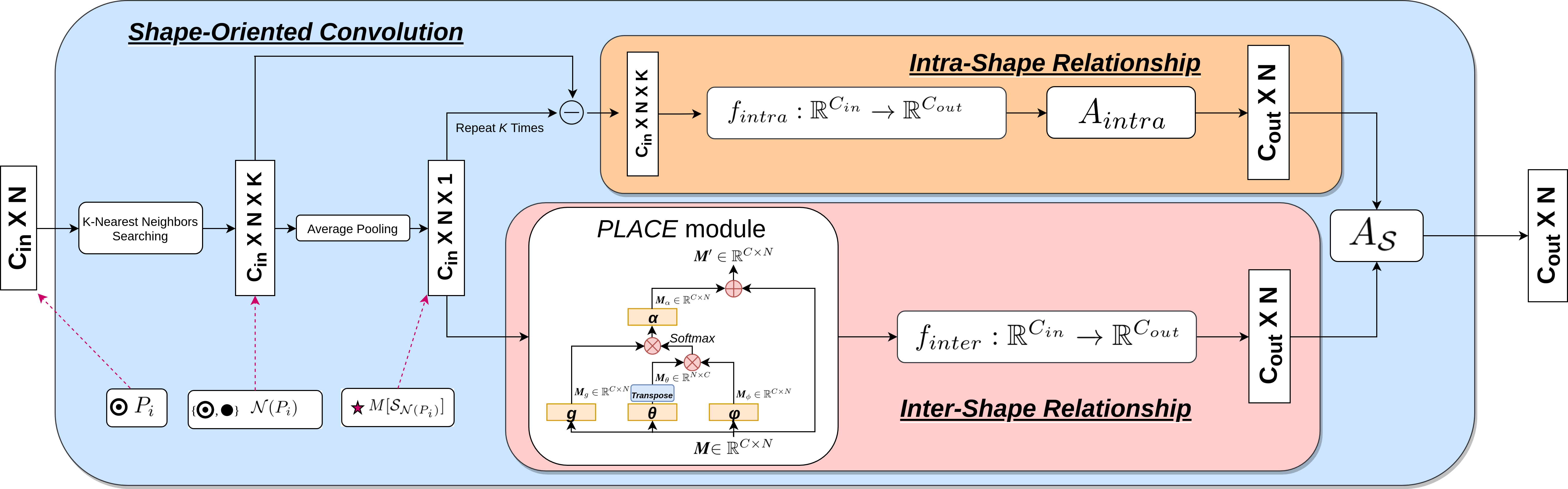} 
\caption{\textit{ShapeConv} operator proposed, modeling two shape-oriented relationships, which are the \textit{intra-shape relationship} and the \textit{inter-shape relationship}. $A_{\mathcal{S}}$ is the shape-oriented aggregation function of these two relationships, while $f_{intra}$ and $f_{inter}$ are the two channel mapping functions for the learning of these two relationships. $\ominus$ denotes the elementwise subtraction between the features of each neighboring point $P_{j} \in \mathcal{N}(P_i)$ and that of the moment point of the local shape they formed as $\mathcal{S}_{\mathcal{N}(P_i)}$. $A_{intra}$ is the intra-shape aggregation function. The details of the $PLACE$ module can be viewed in Fig.~\ref{fig_place}. }
\label{fig_shapeconv}
\end{figure*}

To this end, \textit{ShapeConv} is proposed to model and aggregate these two shape-oriented relationships contained in point cloud, namely, intra-shape relationship and inter-shape relationship. The output of \textit{ShapeConv} for point $P_i$ is thus described as:
\begin{equation}
    X_{i}^{\prime} = A_{\mathcal{S}}(
    L_{\mathcal{S}}(\mathcal{S}_{\mathcal{N}(P_i)}), \ G_{\mathcal{S}}(
    \mathcal{S}_{\mathcal{N}(P_i)},\ \mathcal{S}_{others}
    )
    ),
\end{equation} where  $\mathcal{S}_{others} = \{\mathcal{S}_{\mathcal{N}(P_j)}\}$ for $1\leq j \leq N $ and $i \neq j$, $L_{\mathcal{S}}(\cdot)$ and $G_{\mathcal{S}}(\cdot,\cdot)$ are learning functions of local intra-shape relationship and global inter-shape relationship, and $A_{\mathcal{S}}(\cdot,\cdot)$ is the aggregation function of these shape-oriented relationships. To keep it simple, elementwise-sum operation is adopted as our $A_{\mathcal{S}}(\cdot)$ here. The other two shape-oriented learning functions will be explained and formulated in the following two sections, and their semantic diagrams are shown as Fig.~\ref{fig_inter} and Fig.~\ref{fig_intra}. The overall design of \textit{ShapeConv} is demonstrated in Fig.~\ref{fig_shapeconv}.
\subsubsection{Intra-Shape Relationship}
\begin{figure}[t]
\centering
\includegraphics[width=\columnwidth]{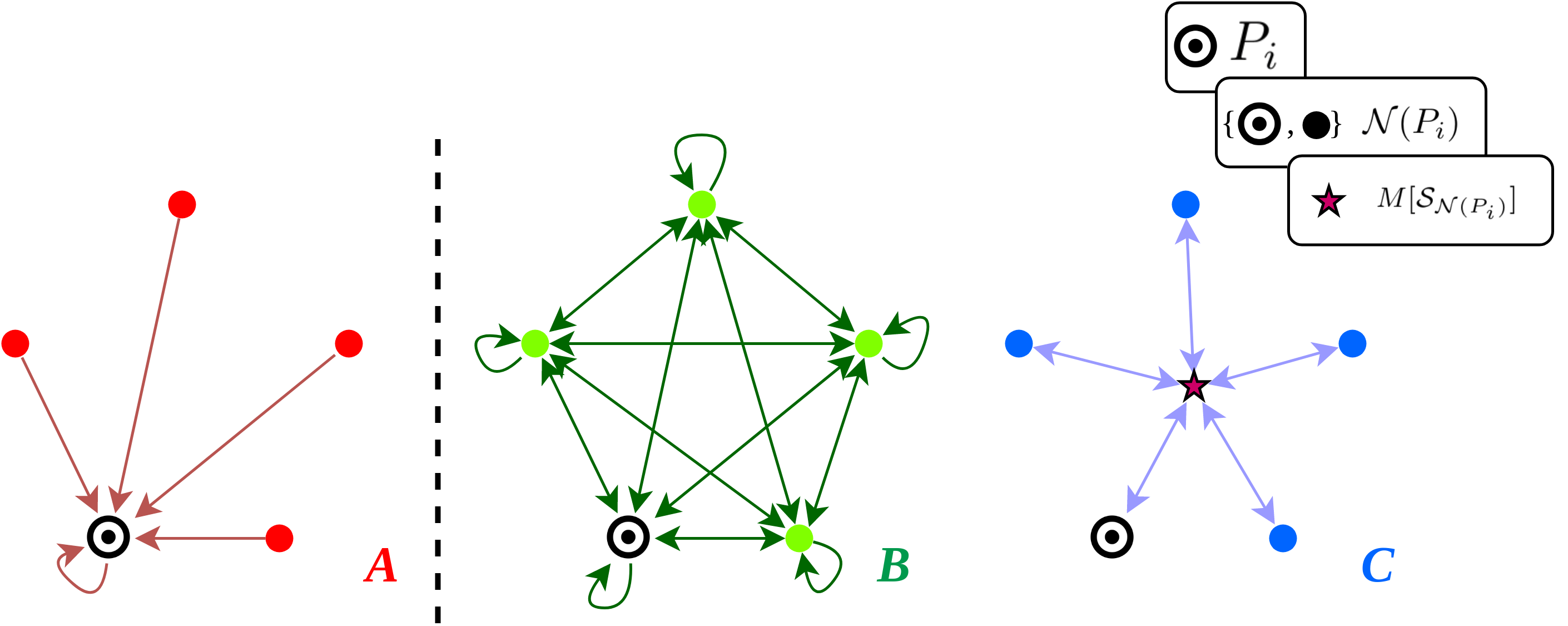} 
\caption{How the pointwise features are aggregated in a local shape $\mathcal{S}_{\mathcal{N}(P_i)}$ for its \textbf{intra-shape relationship} learning, by modeling the pairwise interactions (including self-interaction) between points $P_j \in \mathcal{N}(P_i)$ in different manners: \textit{A}. Only the pairwise interactions targeting at $P_i$ are to be considered, with potential loss of overall geometric information; \textit{B}. All pairwise interactions will be taken into the account via the dense connections built on $\mathcal{N}(P_i)$; \textit{C}. One possible simplified version of \textit{B}, where the moment of shape will be firstly computed as $M[\mathcal{S}_{\mathcal{N}(P_i)}]$ aggregating the overall geometric information, and the interactions between each point and their moment are to be analyzed.}
\label{fig_inter}
\end{figure}

As the local shape $\mathcal{S}_{\mathcal{N}(P_i)}$ is formed by all neighbouring points $P_j \in \mathcal{N}(P_i)$, each neighbouring point $P_j$ is expected to contribute equally to the generation of the complex geometric information locally encoded at $\mathcal{S}_{\mathcal{N}(P_i)}$. This expectation is consistent with the definition of conventional convolution on regular grids, where all pixels within a kernel placed would play the same role for the computation of convolved value, in a ``sum of product" manner.

For this reason, we make $P_j \in \mathcal{N}(P_i)$ densely connected to form $\mathcal{S}_{\mathcal{N}(P_i)}$ (shown in Fig.~\ref{fig_inter} B), rather than treating any point uniquely, such as only building up the connections centralized at $P_i$ among $\mathcal{N}(P_i)$ like in most of the previous works (shown as Fig.~\ref{fig_inter} A). That is, for any point $P_a$ forming our densely-connected shape $\mathcal{S}_{\mathcal{N}(P_i)}$, we can compute its aggregated features $E_a$ representing the pairwise interactions centralized at $P_a$ as:
\begin{equation}
    E_{a} = \frac{1}{N} \underset{P_b \in \mathcal{N}(P_i)}{\Sigma} g(X_a, \ X_b),
\end{equation} where $g(\cdot, \cdot)$ is a pairwise function and it can be upgraded to represent the directed pairwise interactions targeting at $P_a$ with the subtraction implementation, which has been experimentally proven to be more efficient than other similar implementations, such as sum and concatenation~\cite{pointweb}. Interestingly, these aggregated features $E_a$ can therefore be simplified as: 
\begin{multline}
        E_{a} = \frac{1}{N} \underset{P_b \in \mathcal{N}(P_i)}{\Sigma} (X_a - X_b) = X_a - X_{M[\mathcal{S}_{\mathcal{N}(P_i)}]}, \text{ and} \\
        X_{M[\mathcal{S}_{\mathcal{N}(P_i)}]} =  \frac{1}{N} \underset{P_b \in \mathcal{N}(P_i)} {\Sigma} X_b,
\end{multline}
where $M[\cdot]$ geometrically denotes the moment point of local shape and $X_{M[\mathcal{S}_{\mathcal{N}(P_i)}]}$ is the feature of this moment point, which is computed as the averaged feature over $P_j \in \mathcal{N}(P_i)$. Therefore, $E_a$ can be seen as an interaction between $P_a$ and the entire local shape $\mathcal{S}_{\mathcal{N}(P_i)}$, which is demonstrated in Fig.~\ref{fig_inter} C. This formulation can also be seen as Context Normalization~\cite{cn} performed on each dynamically constructed local shape, with the division step excluded.

Furthermore, the intra-shape relationship among $\mathcal{S}_{\mathcal{N}(P_i)}$ can be defined as: 
\begin{equation}
    L_{\mathcal{S}}(\mathcal{S}_{\mathcal{N}(P_i)}) = \underset{P{a}\in\mathcal{N}(P_i)}  {A_{intra}}(f_{intra}(E_a)), 
\end{equation}
where intra-shape aggregation function $A_{intra}$ is supposed to be a symmetry function to achieve the permutation invariance required by unordered point cloud data~\cite{pointnet}, and $f_{intra}: \mathbb{R}^{C_{in}} \xrightarrow{} \mathbb{R}^{C_{out}}$~ is a channel mapping function.

\subsubsection{Inter-Shape Relationship}
\begin{figure}[!t]
\centering
\includegraphics[width=\columnwidth]{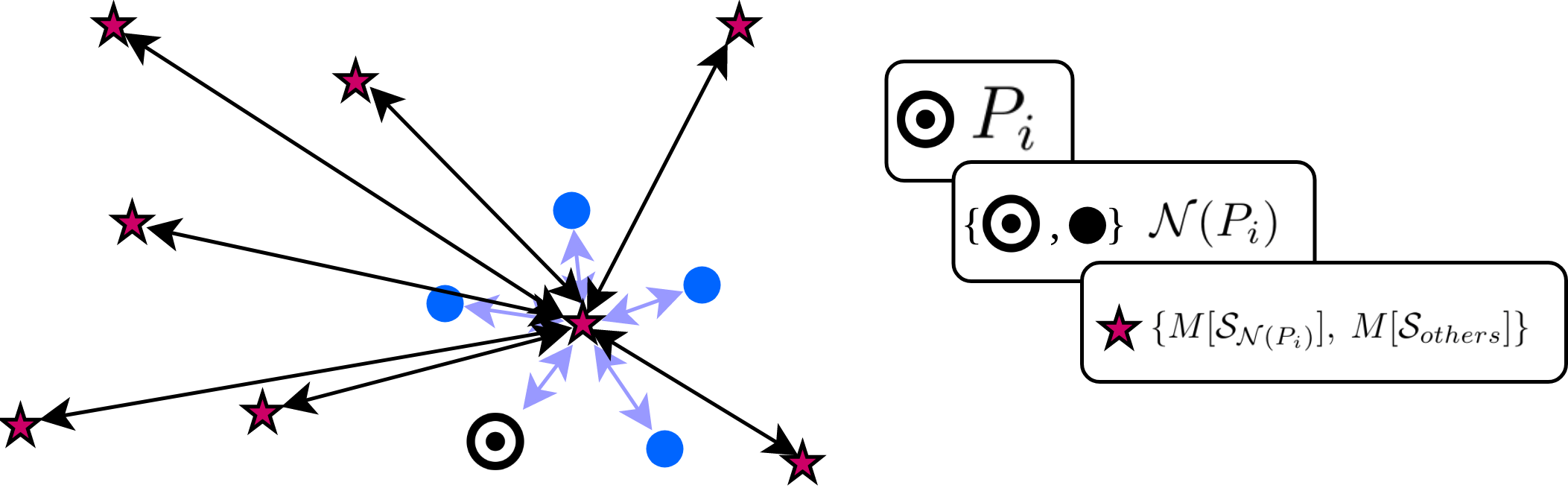} 
\caption{Our \textit{ShapeConv} proposed is also designed to learn the \textbf{inter-shape relationship}, via capturing the long-ranged dependencies between different local underlying shapes.}
\label{fig_intra}
\end{figure}

To learn the inter-shape relationship in point cloud, as demonstrated in Fig.~\ref{fig_intra}, the long-ranged context between each local underlying shape $\mathcal{S}_{\mathcal{N}(P_i)}$ should be taken into the consideration. Inspired by Non-Local Neural Networks~\cite{nonlocal}, several attention-involved modules ~\cite{sagan,danet} have been proposed to learn the long-ranged dependency in the computer vision domain, which are mainly focused on the conventional visual data, such as images and videos. Similar to~\cite{shapecontextattn}, we therefore modify and extend their works to our pointwise version, namely, \textit{pointwise long-ranged attentional context enhancement (PLACE)} module.

\begin{figure}[!b]
\centering
\includegraphics[width=\columnwidth]{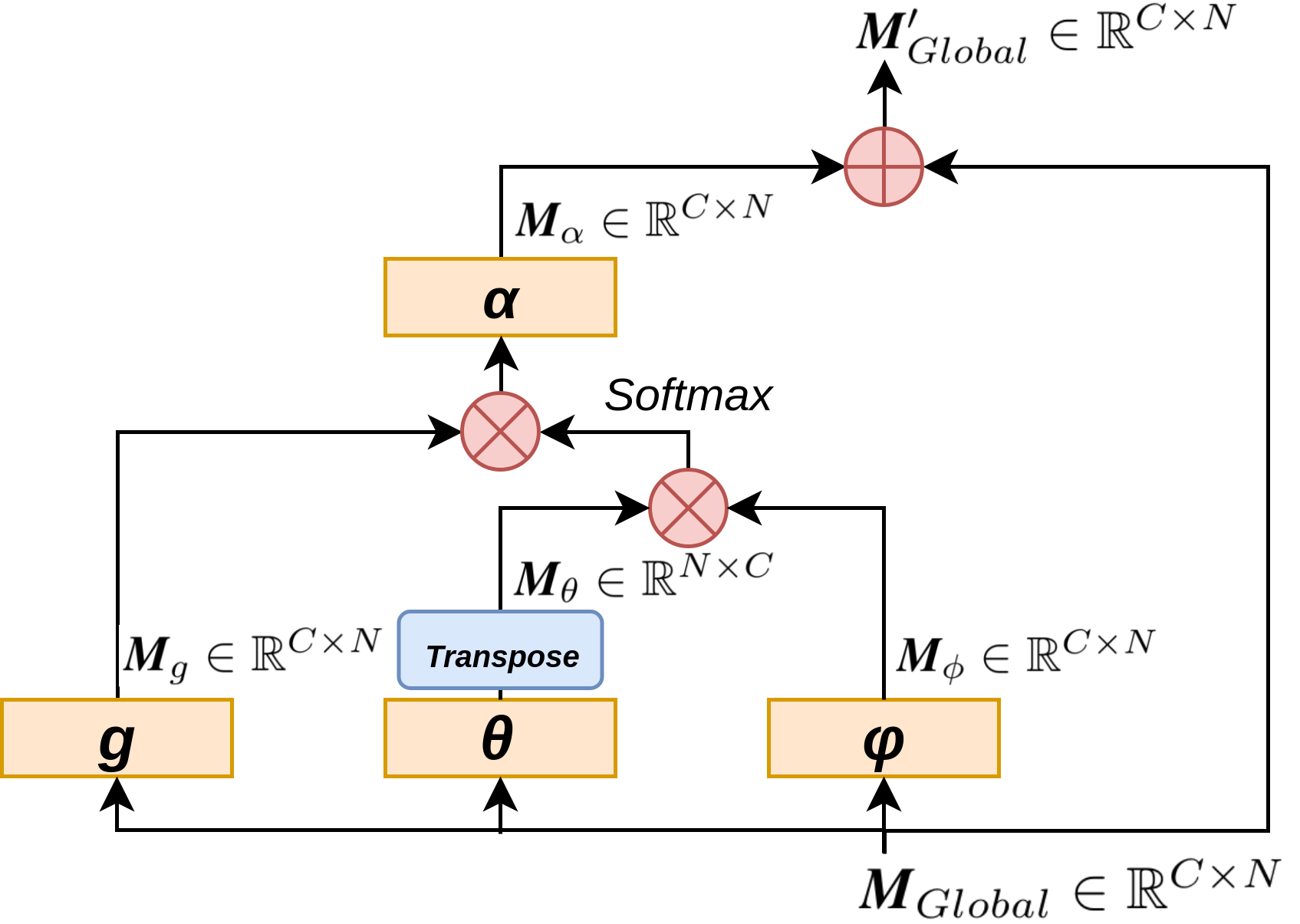} 
\caption{Our \textit{PLACE} module for long-ranged dependency modeling in point cloud, which is simplified from Non-Local Block~\cite{nonlocal}.}
\label{fig_place}
\end{figure}
As clearly illustrated in Fig.~\ref{fig_place}, we first obtain the moment features of all local shapes and group them as a global shape feature matrix $M_{Global} \in \mathbb{R}^{C \times N}$. Then we implement $g$, $\theta$, $\phi$, and $\alpha$ as four separate $conv1d$ with kernel size = 1. Finally, we achieve the global feature enhancement by using the $PLACE$ module to model their long-ranged dependencies, as:
\begin{equation}
    M_{Global}^{\prime} = \textit{PLACE}_{g, \theta, \phi, \alpha}(M_{Global}).
\end{equation} Our inter-shape relationship learning can thus be defined as:
\begin{equation}
    G_{\mathcal{S}}(
    \mathcal{S}_{\mathcal{N}(P_i)},\ \mathcal{S}_{others}
    ) = f_{inter}(M^{\prime}_{Global \ i}),
\end{equation} where $M_{Global \ i}^{\prime}$ is the enhanced global shape feature for $ \mathcal{S}_{\mathcal{N}(P_i)}$ and $f_{inter}: \mathbb{R}^{C_{in}} \xrightarrow{} \mathbb{R}^{C_{out}}$ is a channel mapping function. 

\subsubsection{Design of \textit{ShapeConv} Module.}

We dynamically select k-nearest neighbors around a sampled point $P_i$ to form its local underlying shape $\mathcal{S}_{\mathcal{N}(P_i)}$, where $k=16$ in our implementation. Then an average pooling is applied to compute the averaged feature of the neighboring points $P_j \in \mathcal{N}(P_i)$, which geometrically represents the moment point of local shape $\mathcal{S}_{\mathcal{N}(P_i)}$. The pointwise interaction between each neighboring point and their moment point is calculated as the feature difference between $X_{j}$ and $X_{M[\mathcal{S}_{\mathcal{N}(P_i)}]}$ and be further taken as inputs to learn the intra-shape relationship. Meanwhile, the features of all moment points formed by the local underlying shapes are grouped together and fed into our proposed \textit{PLACE} module for the learning of the inter-shape relationship. Within our proposed \textit{ShapeConv} module, $f_{intra}$ and $f_{inter}$ are two channel mapping functions designed, which can be approximated by multi-layer perceptron (MLP)~\cite{mlp}. Max-pooling is chosen as the symmetric intra-shape aggregation function for $A_{intra}$, and $A_{\mathcal{S}}$ is implemented using the elementwise sum operation to incorporate the pointwise features from global context and local context.

\subsection{SOCNN: Shape-Oriented Convolutional Neural Network}

\begin{figure*}
\centering
\includegraphics[width=\textwidth]{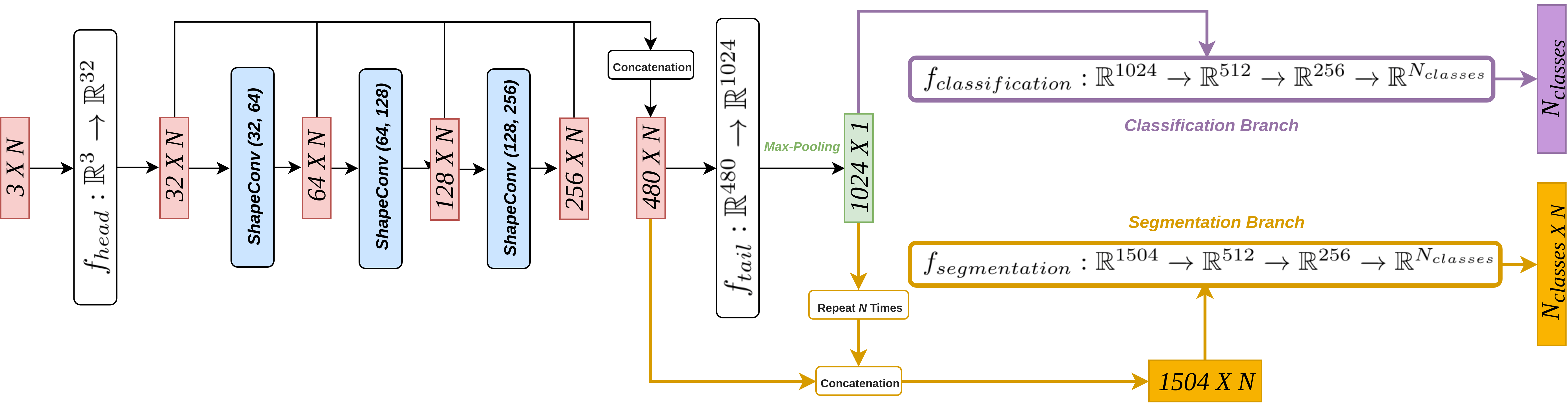} 
\caption{Shape-Oriented Convolutional Neural Network proposed, containing the \textit{classification branch} and the \textit{segmentation branch}. $N$ is the number of sampled points. $ShapeConv(m, n)$ represents the module demonstrated in Fig.~\ref{fig_shapeconv}, with $C_{in} = m$ and $C_{out} = n$. $f_{head}$, $f_{tail}$, $f_{classification}$, and $f_{segmentation}$ are the channel mapping functions applied. }
\label{fig_socnn}
\end{figure*}

As illustrated in Fig.~\ref{fig_socnn}, there are three consecutive \textit{ShapeConv} modules stacked in our \textit{SOCNN} architecture, to capture the intra-shape relationship and inter-shape relationship encoded in point cloud data. Then the multi-scale features from these \textit{ShapeConv} modules are combined through the shortcut connections, while a global shape signature of this point cloud object is further symmetrically aggregated using max-pooling~\cite{pointnet}. $f_{head}$ and $f_{tail}$ are the two channel-raising mapping functions, which are approximated by MLP.

The classification branch is implemented by another channel mapping function $f_{classification}$ applied on the global shape signature computed, while the segmentation branch would output the per-point classifications, by taking both of the learned multi-level representations and the global shape signature into the consideration. The two extra channel mapping functions $f_{classification}$ and $f_{segmentation}$ are implemented by MLP as well.

\section{Experiment}
\subsection{Implementation Details}
We select Adam as the optimizer, with learning rate 0.001 and cosine annealing applied~\cite{cosine}. Batch size is set to 32, while the momentum of batch normalization is initially set as 0.9 and decays with a rate of 0.5 for every 30 epochs. BatchNorm and LeakyRelu are used in all layers and omitted in figures above for simplification purpose. Dropout layers (with dropout rate = 0.5) are adopted within $f_{classification}$. The overall training framework is implemented on Pytorch with two NVIDIA GTX 1080Ti GPUs, using a distributed training scheme with Synchronized BatchNorm proposed~\cite{syncbn}.

\subsection{Shape Classification Task}
We firstly evaluate our model on the ModelNet40 dataset~\cite{modelnet40} for point cloud classification task. This dataset consists of 9843 training 3D models and 2468 testing models, which are collected for 40 shape categories. We follow the same experimental setting used by \cite{pointnet}. For each raw 3D model from ModelNet40, we discard the mesh data after generating their corresponding point cloud data, by uniformly sampling 1024 points with (x, y, z) coordinates as their initial pointwise features. During the point sampling processing, the meshes data are discarded and their (x, y, z) coordinates are normalized to re-scale the 3D objects into unit spheres. The \textit{classification branch} of our \textit{SOCNN} is used for this shape classification task. Simple point cloud data augmentation techniques are adopted on the raw point coordinates, which are random scaling, translation, and perturbing. Similar to~\cite{pointnet,pointnet2,rscnn}, ten voting tests are applied for each testing instance and their averaged results are computed as the final predictions.

Compared with other state-of-the-art approaches, our \textit{SOCNN} achieves comparably significant results for the task of point cloud classification, which is demonstrated in Table~\ref{table_modelnet40}. To the best of our knowledge, among all the methods manipulating point cloud data directly, \textit{SOCNN} is the first method proposed to incorporate the global context and local context, by modeling the two shape-oriented relationships independently, i.e., the \textit{intra-shape relationship} and the \textit{inter-shape relationship}. As one of the other top-ranked methods, RS-CNN did not consider the global inference, which is captured and processed by our inter-shape module. Compared to PointWeb, which requires normal vectors calculated from the object meshes and extensively modeled pointwise interactions via a learning-based approach, our design achieves a similar effect in encouraging the information exchange within each local neighborhood, but in a more efficient manner.  

\begin{table}
\centering
\caption{Shape classification results (\%) on ModelNet40 dataset. \textbf{*} denotes additional points sampled for the classification task.}
\begin{tabular}{| l | c| c|}
\hline
Method                              &\#points  & Acc. \\
\hline
ECC (\citeyear{ecc})                &  1k        &  87.4    \\
PointNet (\citeyear{pointnet})      &  1k        &  89.2    \\
PointNet++ (\citeyear{pointnet2})   &  1k        &  90.7    \\
PointNet++* (\citeyear{pointnet2})   &  5k        &  91.9    \\
KD-Net (\citeyear{kdnet})        &  1k        &  90.6    \\
KD-Net* (\citeyear{kdnet})        &  5k        &  91.8    \\
SpiderCNN* (\citeyear{SpiderCNN})    &  5k        &  92.4    \\
SO-Net* (\citeyear{SONET})           &  2k        &  90.9    \\
PCNN (\citeyear{PCNN})             &  1k        &  92.3    \\
DGCNN (\citeyear{dgcnn})            &  1k        &  92.2    \\
DGCNN* (\citeyear{dgcnn})            &  2k        &  93.5    \\
PointWeb (\citeyear{pointweb})      &  1k        &  92.3    \\
RS-CNN (\citeyear{rscnn})           &  1k        &  93.6    \\
RS-CNN* (\citeyear{rscnn})           &  2k        &  93.6    \\
\hline
Proposed                                &  1k         & 93.1     \\
Proposed*         &  2k         &  93.6     \\
\hline
\end{tabular}
\label{table_modelnet40}
\end{table}

\begin{table*}[!htbp]

\centering
\caption{Part segmentation results on ShapeNet-Part dataset. Metric is mIoU(\%) on points.}
\resizebox{\textwidth}{!}{ 
\begin{tabular}{
|           l |     c|   c     c     c     c      c      c        c       c       c       c       c      c        c     c         c        c|}
\hline
              &mean&aero&bag&cap&car&chair&ear&guitar&knife&lamp&laptop&motor&mog&pistol&rocket&skate&table\\
              &      &      &     &     &     &       &phone&        &       &      &        &       &     &        &        &board&       \\
\hline
\# Shapes     &      & 2690 &  76 &  55 & 898 &3758   &69     & 787    &  392  & 1547 & 451    &  202  & 184 &   283  & 66     & 152   & 5271   \\
\hline
PointNet (\citeyear{pointnet})      &83.7  &83.4  &78.7 &82.5 &74.9 &89.6   &73.0   &91.5    &85.9   &80.8  &95.3    &65.2   &93.0 &81.2    &57.9    &72.8   &80.6  \\
PointNet++ (\citeyear{pointnet2})   &85.1  &82.4  &79.0 &87.7 &77.3 &90.8   &71.8   &91.0    &85.9   &83.7  &95.3    &71.6   &94.1 &81.3    &58.7    &76.4   &82.6  \\
PointCNN (\citeyear{pointCNN})    &86.1  &84.1  &\textbf{86.5} &86.0 &\textbf{80.8}&90.6   &79.7   &\textbf{92.3}   &\textbf{88.4}  &85.3  &96.1   &\textbf{77.2 } &\textbf{95.3} &84.2    &\textbf{64.2}    &80.0   &83.0  \\
SpiderCNN (\citeyear{SpiderCNN})     &85.3  &83.5  &81.0 &87.2 &77.5 &90.7   &76.8   &91.1    &87.3   &83.3  &95.8    &70.2   &93.5 &82.7    &59.7    &75.8   &82.8  \\
PCNN (\citeyear{PCNN})      &85.1  &82.4  &80.1 &85.5 &79.5 &90.8   &73.2   &91.3    &86.0   &85.0  &95.7    &73.2   &94.8 &83.3    &51.0    &75.0   &81.8  \\
KCNet (\citeyear{KCNet})        &84.7  &82.8  &81.5 &86.4 &77.6 &90.3   &76.8   &91.0    &87.2   &84.5  &95.5    &69.2   &94.4 &81.6    &60.1    &75.2   &81.3  \\
DGCNN (\citeyear{dgcnn})       &85.1  &\textbf{84.2} &83.7 &84.4 &77.1 &90.9   &78.5   &91.5    &87.3   &82.9  &96.0    &67.8   &93.3 &82.6    &59.7    &75.5   &82.0  \\
RS-CNN (\citeyear{rscnn})       &\textbf{86.2} &83.5  &84.8 &\textbf{88.8}&79.6 &91.2  &\textbf{81.1}  &91.6    &\textbf{88.4}  &\textbf{86.0} &96.0    &73.7   &94.1 &\textbf{83.4}   &60.5   &\textbf{77.7}  &\textbf{83.6}  \\
\hline
Proposed          &  85.7    & 83.9   & 84.1  & 85.0  & 77.4 &\textbf{91.3}   &  78.3  &  91.7 & 87.4  &  83.8   & \textbf{96.4}   & 69.7      &  93.5   &  83.1      &     58.9   &    76.2   &  82.9 \\
\hline
\end{tabular}
}
\label{table_segmentation}
\end{table*}

\subsection{Shape Part Segmentation Task}
We then further evaluate our model on the ShapeNet-Part dataset~\cite{shapenet} for the point cloud segmentation task. This dataset consists of 16881 3D objects, covering 16 shape categories. Most of the point cloud instances are annotated with less than six part labels, and there exist 50 parts labels in total. We split dataset into 12137 training objects, 1870 validation objects, and 2874 testing objects, following the official split policy announced by~\cite{chang2015shapenet}. For each 3D shape object, its corresponding point cloud data is generated by 2048 points sampled uniformly with (x, y, z) coordinates as their initial pointwise features. The \textit{segmentation branch} of \textit{SOCNN} is used for this point cloud segmentation task.

\begin{figure}[!b]
\centering
\includegraphics[width=\columnwidth]{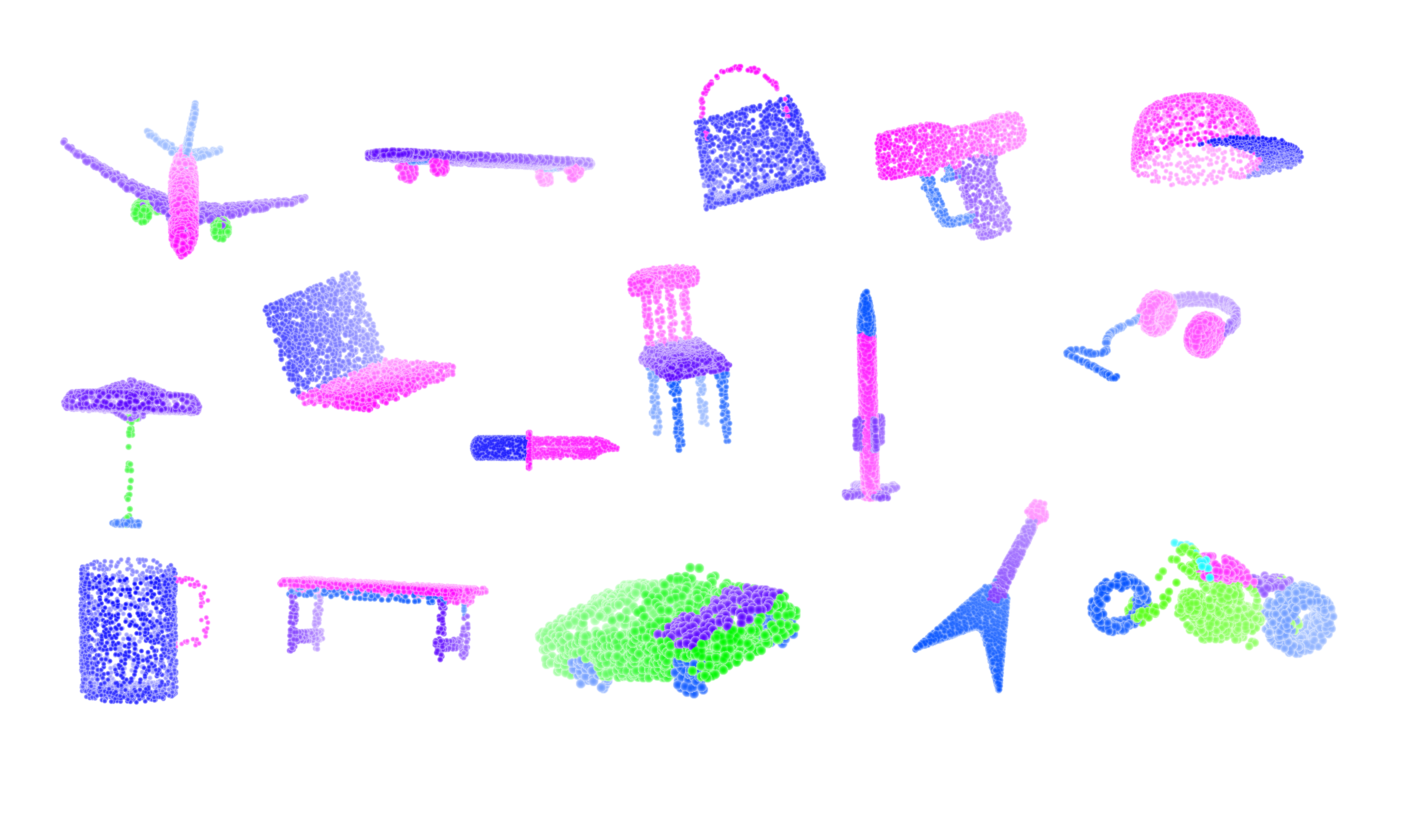} 
\caption{Part segmentation examples on the ShapeNet-Part dataset.}
\label{fig_shapenet_demo}
\end{figure}

Following the same evaluation metrics set by PointNet~\cite{pointnet}, we calculate the Intersection-over-Union (IoU) of our point cloud part segmentation results. Specifically, the comparisons are made in terms of per-object-category IoUs and the mean IoU (mIoU). To make a fair comparison, we evaluate our model with other state-of-the-arts approaches, which were proposed to manipulate point cloud data directly and would sample 2048 points for each object for the part segmentation task. The visual outputs generated by our \textit{SOCNN} proposed for the part segmentation task are shown in Fig.~\ref{fig_shapenet_demo}. As presented in Table~\ref{table_segmentation}, the quantitative comparisons demonstrate that our model achieves state-of-the-art performance on the part segmentation task of the \textit{CHAIR} objects and the \textit{LAPTOP} objects.

\subsection{Ablation Studies}
We design and perform extensive ablation studies on ModelNet40 dataset to analyze the significance of different components proposed for the shape-oriented relationship modeling. The results of the ablation studies can be viewed in Table~\ref{table_abs}. Models A and B are implemented by manipulating their pointwise features directly, rather than computing the feature difference between each point and the moment point of local shapes. To retain their number of parameters and thus make a fair comparison with other models, the neighbouring features and source feature from each neighbourhood $\mathcal{N}(P_i)$ are selected for the intra- and inter-shape relationship modeling for the formed $\mathcal{S}_{\mathcal{N}(P_i)}$, respectively. It can be seen from the results that the modeling of both the intra-shape relationship and inter-shape relationship has positive influences towards the final classification. Compared to the inter-shape relationship, the intra-shape relationship may contribute more to the final recognition results. Notably, the calculation of moment points itself can be understood as a quick and efficient operation to aggregate the local context and global context, which enhances the performance of models from C to E significantly.
\begin{table}
\centering
\caption{Ablation studies designed for \textit{SOCNN} (\%). MP denotes whether the model calculates moment points and uses them for the relationship learning. INTRA and INTER indicate whether the model contains intra-shape relationship modeling and inter-shape relationship modeling, respectively. }
\begin{tabular}{| l | c| c | c  c | c|}
\hline
model  & \#points & MP & INTRA & INTER & Acc. \\
\hline
A     &  1k      &               & \checkmark  &              &  89.5 \\
B     &  1k      &               &             & \checkmark   &  88.2 \\
\hline
C     &  1k      & \checkmark    & \checkmark  &              &  90.9 \\
D     &  1k      & \checkmark    &             &  \checkmark  &  88.7 \\
E     &  1k      & \checkmark    & \checkmark  &  \checkmark  &  93.1 \\
F     &  2k      & \checkmark    & \checkmark  &  \checkmark  &  93.6 \\
\hline
\end{tabular}
\label{table_abs}
\end{table}

\section{Conclusion}
In this paper, we firstly define two shape-oriented relationships existing in point cloud data and reformulate their geometric representation learning as two modeling processes for the global context information and the local context information. Unlike previous geometric information modeling performed for point clouds, the shape-oriented convolution (\textit{ShapeConv}) module is proposed to incorporate the contextual effects caused by the intra-shape relationship and inter-shape relationship and aggregate these effects to update the pointwise features. Notably, we experimentally observe that the computation of the moment point from a local underlying shape can be seen as a simple but efficient way to combine the contextual information captured at both the global level and local level. Finally, we propose the shape-oriented convolution neural network (\textit{SOCNN}) for point cloud analysis and evaluate its significance in the point cloud tasks of shape classification and shape part segmentation.

\bibliographystyle{aaai}
\bibliography{4075-ref}

\end{document}